%% file: ycbev.tex
\documentclass[runningheads]{llncs}

\usepackage{eccv}

\usepackage[T1]{fontenc}

\usepackage[pagebackref,breaklinks,colorlinks]{hyperref}

\usepackage{times}
\usepackage{enumitem}

\usepackage[utf8]{inputenc}
\usepackage[english]{babel}
\usepackage{amsmath}
\usepackage{amsfonts}
\usepackage{amssymb}
\usepackage{graphicx}
\usepackage{bm} 

\usepackage[dvipsnames]{xcolor}

\usepackage{caption}

\usepackage{wrapfig} 

\usepackage{float}

\usepackage{pifont}
\newcommand{\xmark}{\color{purple} \ding{55}}
\newcommand{\cmark}{\color{olive} \ding{51}}

\usepackage{tablefootnote}

\title{YCB-Ev 1.1: Event-vision dataset for 6DoF object pose estimation}
\author{Pavel Rojtberg\orcidID{0000-0001-9736-6866} \and Thomas Pöllabauer\orcidID{0000-0003-0075-1181}}

\makeatletter

\hypersetup{
    pdftitle={\@title}
}

\begin{document}
	    \institute{Fraunhofer IGD, Darmstadt, Germany \email\{pavel.rojtberg, thomas.poellabauer\}@igd.fraunhofer.de}
	\maketitle
\begin{abstract}
    Our work introduces the YCB-Ev dataset, which contains synchronized RGB-D frames and event data that enables evaluating 6DoF object pose estimation algorithms using these modalities.
    This dataset provides ground truth 6DoF object poses for the same 21 YCB objects \cite{calli2017yale} that were used in the YCB-Video (YCB-V) dataset \cite{xiang2018posecnn}, allowing for cross-dataset algorithm performance evaluation.
    The dataset consists of 21 synchronized event and RGB-D sequences, totalling 13,851 frames (7 minutes and 43 seconds of event data). Notably, 12 of these sequences feature the same object arrangement as the YCB-V subset used in the BOP challenge \cite{sundermeyer2023bop}.
    Ground truth poses are generated by detecting objects in the RGB-D frames, interpolating the poses to align with the event timestamps, and then transferring them to the event coordinate frame using extrinsic calibration.
    Our dataset is the first to provide ground truth 6DoF pose data for event streams. Furthermore, we evaluate the generalization capabilities of two state-of-the-art algorithms, which were pre-trained for the BOP challenge, using our novel YCB-V sequences.
    The dataset is publicly available at \url{https://github.com/paroj/ycbev}.
    
\end{abstract}

	\begin{figure}
		\centering
		\captionsetup{type=figure}
		\includegraphics[width=0.32\linewidth]{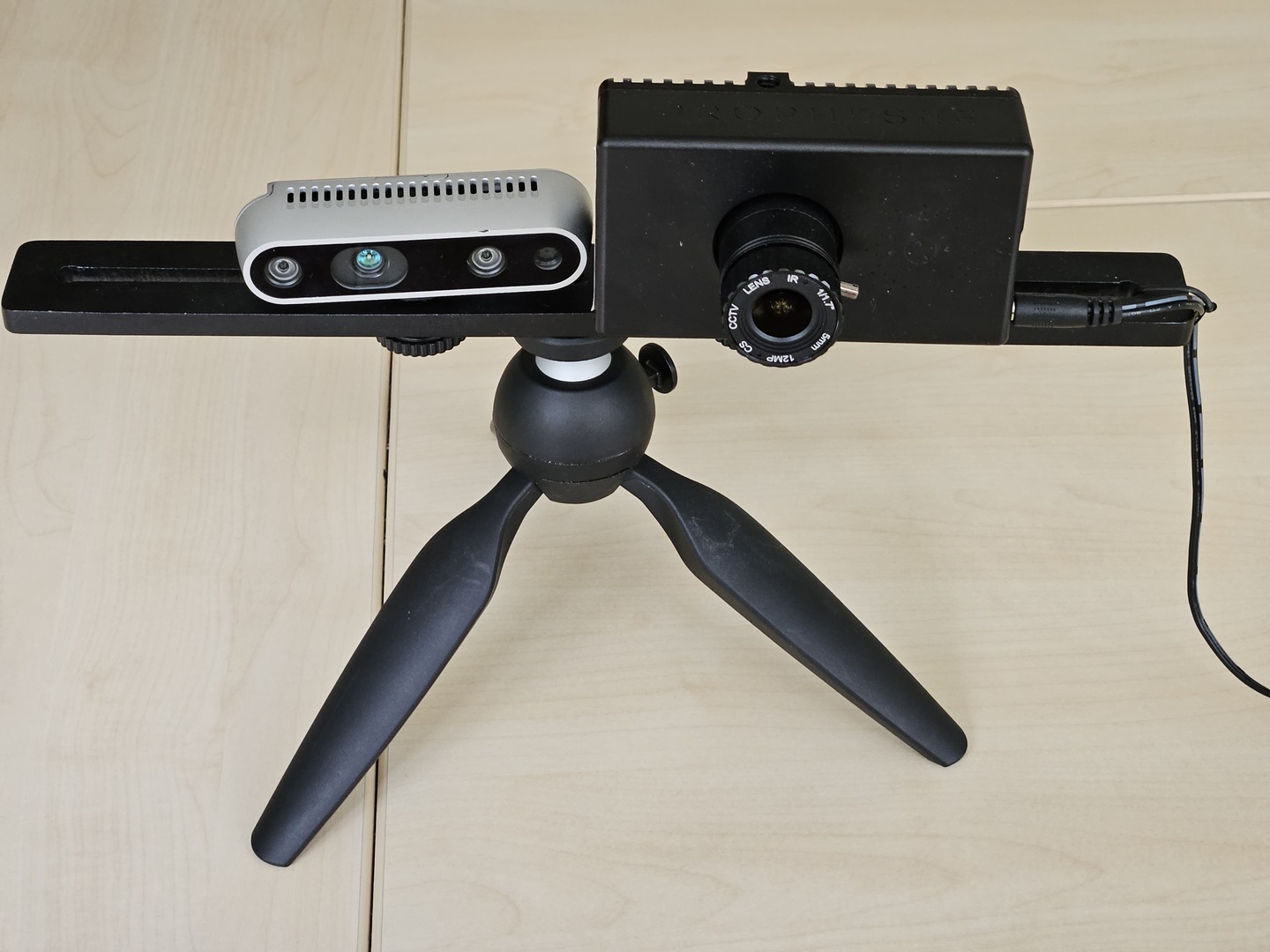}
		\includegraphics[width=0.32\linewidth]{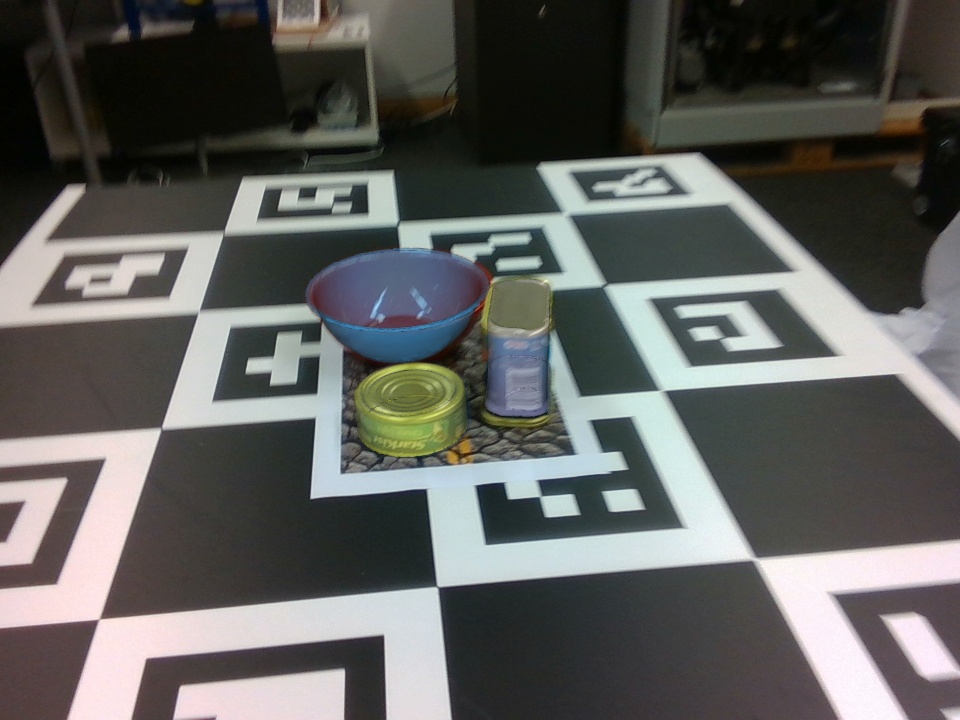}
		\includegraphics[width=0.32\linewidth]{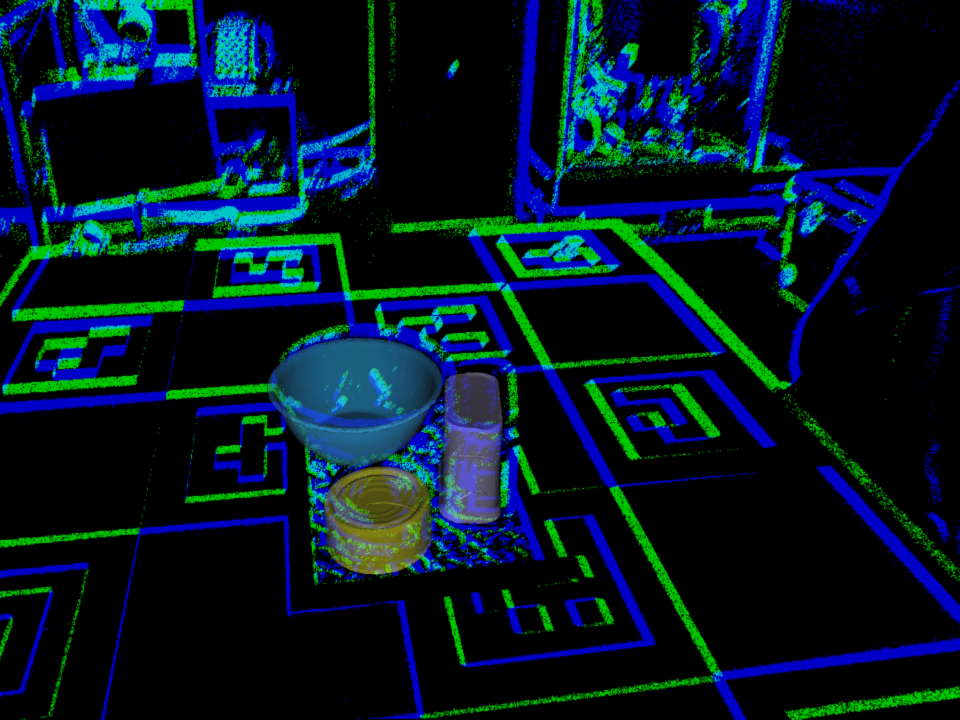}
		\captionof{figure}{The Intel RealSense D435 RGB-D camera (on the left) and the Prophesee EVK2 event camera (on the right) mounted side by side, along with their respective annotated images.
			Throughout this paper, the polarity of the events is represented as blue (falling) and green (rising).}
		\label{fig:teaser}
	\end{figure}%

    \section{Introduction and related work}
    The ability to perform real-time object detection and 6D pose estimation is essential for applications in augmented reality, virtual reality, and robotics.

    The progress in this field is evaluated through the BOP challenge \cite{sundermeyer2023bop}, which ranks algorithms and publishes a leader-board annually.
    To this end, a variety of datasets are utilized, differing in terms of captured modality (e.g., RGB color, depth) and type of objects (e.g., household items, industrial objects). 
    In this context, the algorithms are also evaluated in terms of their ability to generalize from synthetic data to real-world conditions. Since acquiring ground truth data for 6D poses is challenging, many algorithms are trained on synthetic renderings, where the pose is easily accessible. However, even with the use of a physically-based rendering (PBR) pipeline, a domain-gap exists between the renderings and real images. This gap is a specific type of dataset-bias \cite{torralba2011unbiased}.
    Although the domain-gap is measured in the BOP challenge by evaluating algorithms solely on PBR images, the broader impact of dataset bias may extend to neglecting capturing effects such as camera noise, motion blur, and lighting.

    Event-based or neuromorphic cameras provide a novel capturing modality that offers several benefits over classical frame base cameras, such as high-frequency output, high dynamic-range, and lower power-consumption.
    However, these cameras produce a sparse, asynchronous stream of events, which differs from conventional, dense images.
    Instead of reading the entire sensor at once, individual pixels trigger asynchronously when the brightness difference crosses a threshold, generating events at the pixel location that carry the polarity of the threshold (see Figure \ref{fig:teaser}).
 
    In the past, only small-scale datasets with minimal camera motion were available for event cameras \cite{orchard2015converting,serrano2015poker}.
    Even now, only a limited number of automotive centric datasets \cite{sironi2018hats,de2020large} are available that only consider the task of 2D object detection.
    Although it is possible to convert RGB datasets for pose estimation into events using the vid2e tool \cite{Gehrig_2020_CVPR}, there is no publicly available real-world dataset for this task that we are aware of.
 
    The YCB-Video (YCB-V) dataset \cite{xiang2018posecnn} is a notable choice from the datasets utilized in the BOP challenge as it not only provides 3D data enabling the generation of synthetic renderings, but also offers the opportunity to obtain physical objects from the YCB organizers \cite{calli2017yale}.
    \autoref{tbl:ycbdatasets} provides an overview of other YCB-based datasets. Existing datasets offer real and synthetic color and depth data, but they do not include the event modality.
    Having access to the event data of the YCB objects allows for evaluating object detection beyond the existing, automotive-centric event datasets. Furthermore, the pose data annotations enable exploring new applications of event cameras. It enables pose estimation directly on event data without the need for an additional color camera and facilitates high-speed object tracking, as event cameras can sample at rates well beyond 30Hz.
    
    In this work, we acquired the physical YCB objects and recreated the sequences of the YCB-V dataset. While the original dataset was captured with a Asus Xtion Pro Live RGB-D camera, we used a more modern Intel RealSense D435 camera, which allowed us to capture color and depth at the full FOV at 1280x720 px, 30fps, without the need for cropping.
    Simultaneously, we captured event data using the Prophesee EVK2 camera, which was calibrated to the color camera (see Figure \ref{fig:teaser}).

    Based on the above, our key contributions are;
    \begin{enumerate}[nolistsep]
        \item A real-world event dataset with ground-truth poses and
        \item new YCB-V sequences that we used to evaluate the generalization capabilities of top-performing algorithms from the BOP challenge trained on YCB-V.
    \end{enumerate}
 
    This paper is structured as follows: Section \ref{sec:datacap} describes our data acquisition and labelling pipeline.
    Section \ref{sec:datafmt} explains the structure and storage of the captured data, while Section \ref{sec:evaluation} presents the results of generalization evaluation we performed.

    We conclude with Section \ref{sec:conclusion}, which summarizes our results, discusses the limitations, and outlines potential future work.

	\begin{table*}
	\centering
	\begin{tabular}{|c|c|c|c|c|c|c|}
		\hline
		Dataset & object classes & scenes & high-res $^1$ & real data & challenging & event data \\
		\hline
		\hline
		YCB \cite{calli2017yale} & 77 & 77$^2$  & \cmark & \cmark & \xmark & \xmark \\
		\hline
		Falling things \cite{tremblay2018falling} & 21 & 3075 & \xmark & \xmark & \cmark & \xmark \\
		\hline
		YCB-V \cite{xiang2018posecnn} & 21 & 92 & \xmark & \cmark & \xmark & \xmark \\
		\hline
		YCB-M \cite{grenzdorffer2020ycb} & 20 & 32 & \cmark & \cmark & \xmark & \xmark \\
		\hline
		YCB-Ev (ours) & 21 & 21 & \cmark & \cmark & \cmark & \cmark \\
		\hline
	\end{tabular}
	\caption{Overview of related, YCB-based, datasets. Our dataset stands out as the only one that incorporates the event modality and challenging real image sequences.}
	\footnotesize{$^1$ The image resolution is at least HD (1280x720px).\\
		$^2$ There is only one object per scene in the YCB dataset.}\\
	\label{tbl:ycbdatasets}
	\end{table*}
	
    \section{Data capturing and labelling}
    \label{sec:datacap}
    In this section, we describe our approach for capturing high-quality pose data to annotate the event-camera stream. \autoref{fig:teaser} illustrates our capturing setup.

    We chose the Intel Realsense D435 RGB-D camera over the newer and more capable Microsoft Azure Kinect DK as the former allows passive depth capturing.
    The pattern projected by the IR projector in the active depth capturing mode is picked up by the event camera thus rendering the event stream unusable.
    However, using passive mode leads to reduced depth resolution compared to the YCB-V dataset.
    
    Since there are no established algorithms for pose estimation on event data, we instead generate poses using the RGB data and transfer them to the coordinate frame of the event camera via stereo calibration. The global shutter feature of the Realsense camera provides an additional advantage over the Azure Kinect in this context.

    \subsection{Calibration}
    \label{sec:calibration}

    \begin{figure}
        \includegraphics[width=0.49\textwidth]{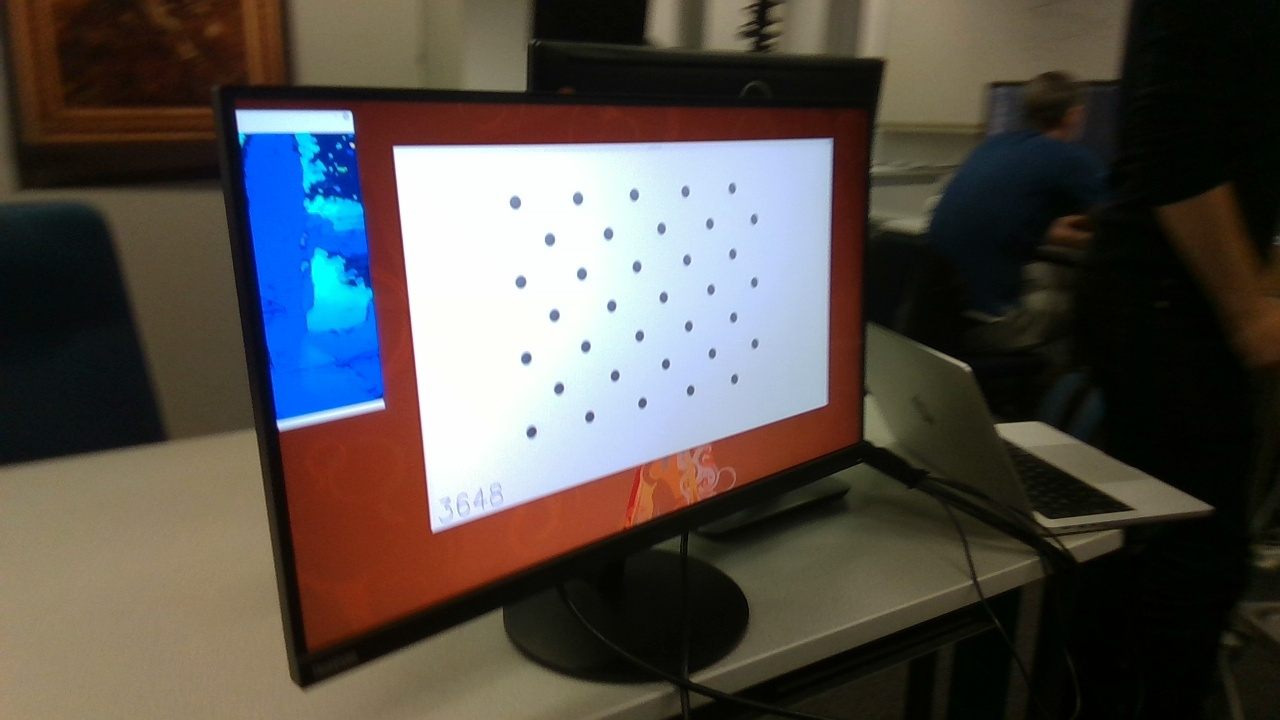}
        \includegraphics[width=0.49\textwidth]{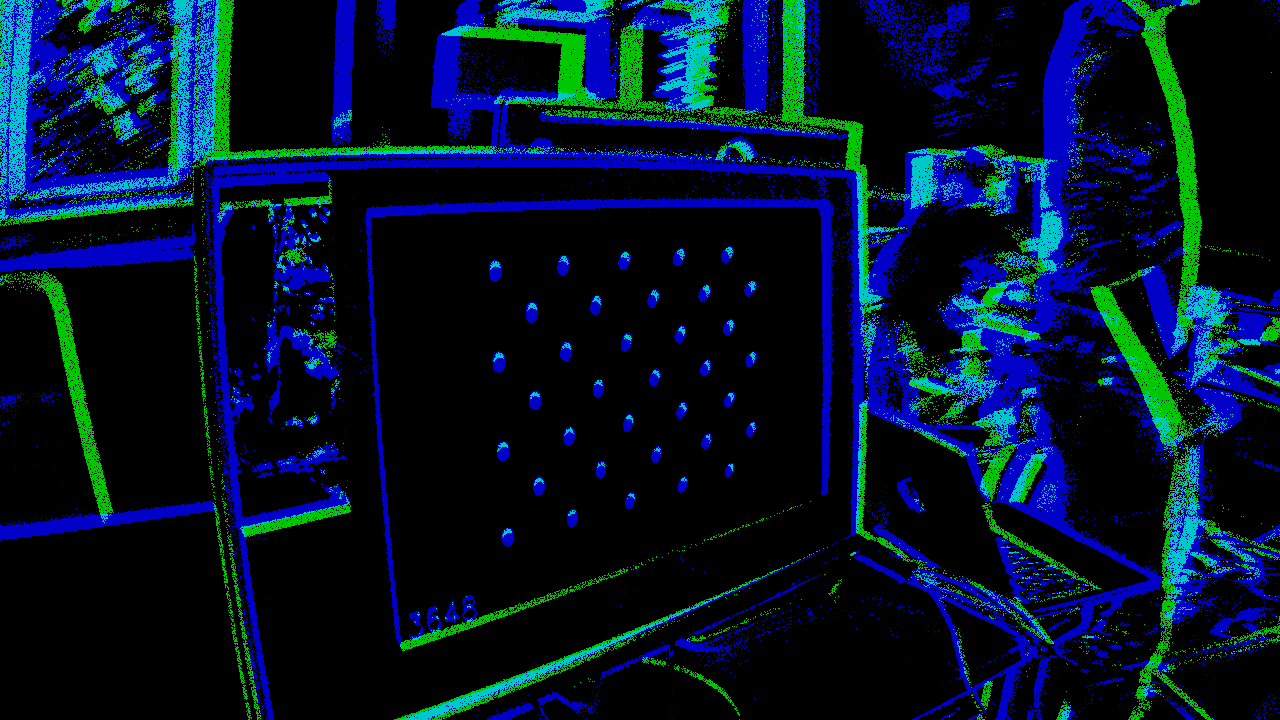}
        \caption{Our method for joint event and RGB camera calibration relies on a flashing blob pattern that can be detected by both sensor technologies.}
        \label{fig:calibration}
    \end{figure}

    \autoref{fig:calibration} depicts the setup employed to obtain calibration data for both the intrinsic calibration of the individual cameras and the extrinsic calibration needed for stereo alignment. We followed the approach described in \cite{mueggler2014event} by using a flashing blob pattern displayed on a screen. Utilizing a screen instead of a printed pattern offers precision advantages, even when calibrating conventional RGB cameras.
    
	We adopted the calibration approach outlined by \cite{rojtberg2018efficient} to ensure reliable calibration data acquisition.

    \subsection{Pose annotations}
    \label{sec:poseann}
    
	To generate ground truth poses using RGB data, we follow this procedure:

	First, we establish a global coordinate system, which we use to combine camera tracking data with object detections from individual frames. For this purpose, we utilize a fiducial marker board \cite{charuco}, known for its robust and precise detection. Although we also experimented with structure-from-motion algorithms \cite{schoenberger2016sfm, dust3r_cvpr24}, we found that their precision was insufficient for our needs.
    
	To position the objects within the global coordinate system, we use multiple frames to obtain a reliable pose estimate for each object. This pose estimation is carried out using the Cosypose algorithm \cite{labbe2020cosypose}, followed by local refinement with the ICG algorithm \cite{stoiber2022iterative}. The object's composition remains fixed for each sequence, and the pose derived from camera tracking is used to project the objects into the corresponding frame as the ground truth poses.
    
    However, due to the challenging capturing conditions in our sequences, camera tracking can sometimes fail. In such cases, we rely on local tracking of individual objects using the ICG algorithm, initializing it at the last frame where camera tracking was successful.
    To minimize drift and occlusion-related inaccuracies, manually refined keyframes are strategically placed, resetting the ICG tracking.
    
    Despite our efforts to achieve precise poses, several sources of inaccuracy remain in this process. First, the global object composition is prone to imprecision since we do not capture specific keyframes for this purpose, but instead use the standard sequence frames. Additionally, camera tracking accuracy decreases with increased motion blur, directly impacting the pose annotations.
    
    \subsection{RGB and Event data synchronization}
    The color images are captured at fixed time intervals determined by the frame rate. In contrast, the event data stream is continuous and does not have fixed time intervals. This makes it difficult to use simple synchronization techniques, such as using an external hardware trigger, that are commonly used in stereo camera systems.
    
    Instead, we resort to displaying a blinking counter on a screen that can be synchronized to in both colour and event images. The counter was captured at the beginning of each sequence.
    
    The method of using a blinking counter on a screen to synchronize color images with event data has a drawback in that it has a minimum possible latency. The counter's blinking frequency is not relevant as the on and off events provide discrete time points. However, the color camera operates at 30 frames per second, resulting in a 33 ms exposure time. For instance, if the counter turns off at the end of the exposure, it may take 33 ms for the corresponding "off" events to be observed in the RGB frame. In practice, the event data, must be resampled to a fixed window-size for visualization. While the window-size can be as low as 1 ms without any issues, this still limits the synchronization accuracy to 33 ms.

    \section{Dataset structure and usage}
    \label{sec:datafmt}
    This section focuses on the data format and structures employed by the supplementary programs for reading and processing the dataset.

    It is important to note that while we only provide ground truth 6D pose data, additional annotations such as 2D and 3D bounding boxes and per-pixel segmentation can be easily generated from the available 3D meshes by performing rasterization using the provided poses.

    At the top level of our dataset, there are three files that describe the parameters that remain constant throughout the entire dataset:
    \begin{itemize}[noitemsep]
        \item \texttt{calib\_realsense.json} contains the intrinsic calibration for the D435 color camera. The images are undistorted by the camera, hence all distortion coefficients are zero. This is also available in BOP format as \texttt{camera\_d435.json}. Furthermore, the depth intrinsic and the depth to color extrinsic is provided, as the depth images are not aligned.
        \item \texttt{calib\_prophesee.json} contains the intrinsic calibration for the EVK2 camera. You must consider image distortion, for correct pose estimation. This can be accomplished by undistorting the event positions, as demonstrated in the supplementary programs.
        \item \texttt{calib\_stereo\_c2ev.json} contains the extrinsic transformation from the D435 color coordinate frame to the EVK2 coordinate frame.
    \end{itemize}
    
	Each captured sequence is stored in a subfolder that contains the sequence data in the format specified by the BOP dataset. The subfolder contains the following contents:
    \begin{itemize}[noitemsep]
        \item \texttt{rgb} contains the RGB color images saved in JPEG format.
        \item \texttt{depth} contains per-pixel depth in millimeters, saved in 16-bit PNG format, synchronized with the RGB frames by the camera.
        \item \texttt{scene\_gt.json} contains the ground truth poses for the RGB frames.
    \end{itemize}
  
     The event stream that is synchronized with the RGB frames is stored outside the subfolder as \texttt{NN\_events.int32.zst}.
     Due to the lack of a standard format for storing event data, we have developed a custom binary format. Our goal is to create a compact file format that is easy to transmit and accessible on different platforms, especially Python and NumPy. We only store the events without any accompanying metadata. Similar to the prophesee DAT file format \cite{propheseedat}, each event is represented by two 32-bit integer numbers. The first integer stores the timestamp in microseconds, while the second integer stores the packed polarity and x, y coordinates of the event. The integer is arranged in little-endian order, with the first 14 bits storing the x-coordinate, the next 14 bits storing the y-coordinate, and the remaining 4 bits encoding the polarity.
     
    To further minimize file size, we compress the event stream using Zstandard compression \cite{collet2018zstandard}. This method reduces the file size by roughly 60\% while providing fast decompression, enabling the data to be decompressed on-the-fly and stored in its compressed form on disk.

    \subsection{Supplementary programs}
	An example of decoding the Zstandard compressed events and performing pose interpolation to an arbitrary event histogram size can be found in the supplementary NumPy-based program. The program also demonstrates how to obtain an accurate 2D bounding box from the object mesh within the event frame using the provided calibration data.

	Additionally, you will find code for aligning the depth images to the color images and undistorting (rectifying) the event locations.

    \subsection{Aligning ground truth poses to event data}
	The provided ground truth poses are in the frame of reference of the RGB camera. In order to align them with the event camera, the rigid stereo transformation must be applied to the pose data. Additionally, the poses are only provided for discrete time steps, while the event data stream is essentially continuous.

    Typically, events are processed by accumulating them over a certain window-size \cite{gallego2020event}, as a single event contains limited information. In this work, events are depicted in the form of 2D event-frames (see Figure \ref{fig:teaser}).
    However, in this scenario, the pose data is only accurate if the histogram window matches the frame rate of the RGB camera, meaning a 33 ms window size is used. For different window sizes, the poses must be interpolated to the current time step of the event window. Simple linear interpolation can be used for the positions, while spherical linear interpolation is required for the rotations.

     \subsection{Sequences}
    Our dataset consists of 21 sequences, totalling 13,851 frames and a duration of 7 minutes and 43 seconds. These sequences feature diverse object arrangements and lighting scenarios.
    
    \begin{figure}
        \includegraphics[width=0.49\textwidth]{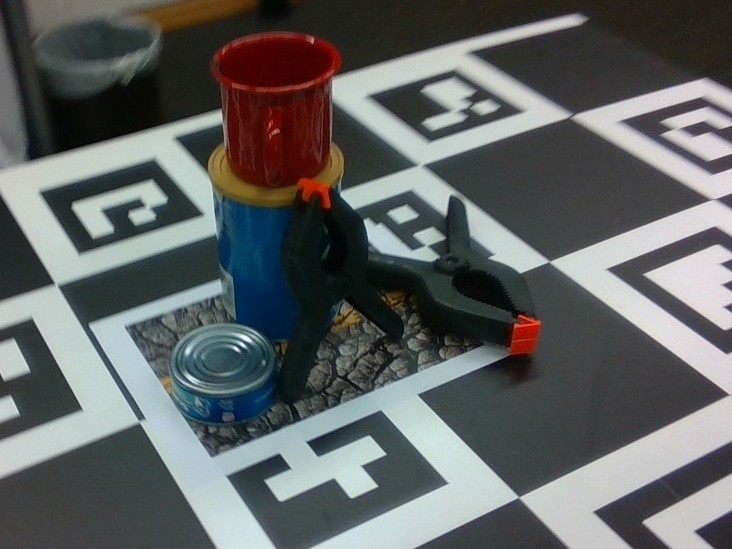}
        \includegraphics[width=0.49\textwidth]{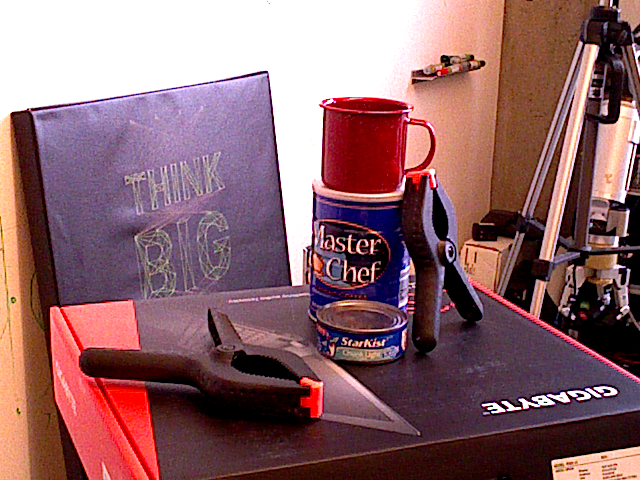}
        \caption{The object arrangement in our dataset (left) corresponds to the object arrangement  in the YCB-V dataset (right)}
        \label{fig:arrangement}
    \end{figure}
    
    The first 12 sequences correspond to objects and arrangements from sequences 48 to 59 in the original YCB-V dataset (see Figure \ref{fig:arrangement}). This particular subset of sequences is also used in the BOP challenge.
    
    \begin{table}
        \centering
        \begin{tabular}{|c|p{0.7\linewidth}|}
            \hline
            Sequence Nr. & Description \\
            \hline
            \hline
            1 -- 12 & Arrangements corresponding to the BOP subset of YCB-V. \\
            \hline
            13 -- 15 & Arragements with more objects, placed at frame border. \\
            \hline
            16, 19 & Additional YCB objects serving as occluders. \\
            \hline
            17 & Seq. 16 arrangement with lights off. \\
            \hline
            18 & Seq. 1 arrangement with lights off. \\
            \hline
            20, 21 & Clumped arrangement with many occlusions. \\
            \hline
        \end{tabular}
        \caption{An overview of the sequence arrangements in our dataset.}
        \label{tbl:sequences}
    \end{table}

    Several of the original YCB objects are no longer available and have been replaced by the YCB organizers. 
    This change affects three objects, "power drill", "pitcher", and "coffee can" from the set of YCB-V objects. This adds an additional challenge of domain adaptation, apart from the different capturing conditions.
    
    \begin{figure}
        \includegraphics[width=0.49\textwidth]{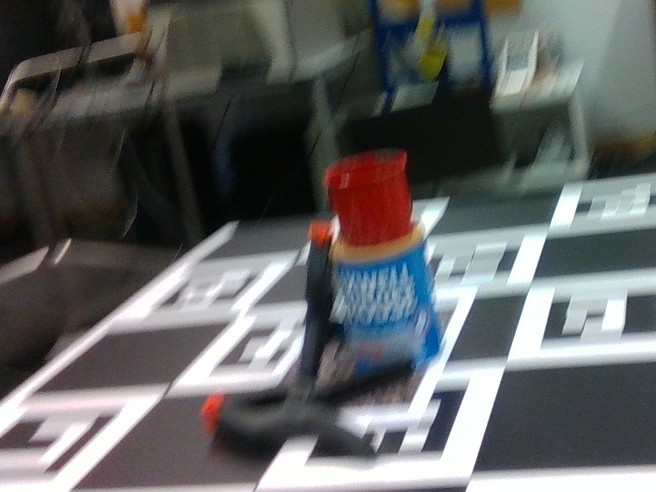}
        \includegraphics[width=0.49\textwidth]{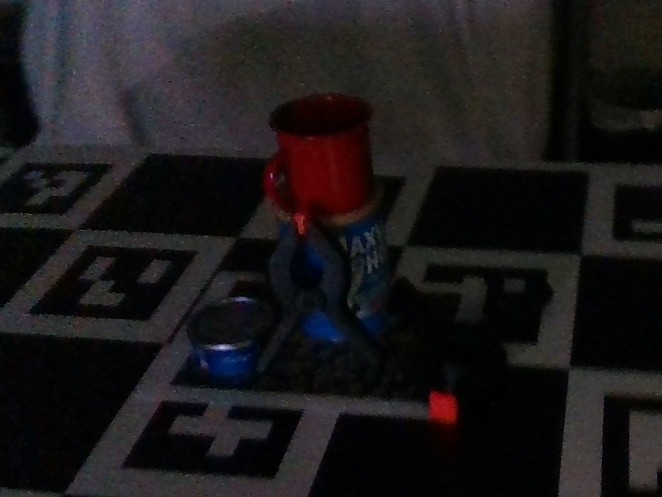}
        \caption{Our dataset contains challenging frames that exhibit fast camera motion and low-light conditions.}
        \label{fig:challenges}
    \end{figure}
    
    In contrast to the YCB-V dataset our sequences also contain fast camera motion and specific sequences are captured under low-light conditions (see Figure \ref{fig:challenges}).
    
    Some sequences in our dataset have challenging capturing conditions, which means that not every frame contains valid ground truth poses for all visible objects. Therefore, frame and object whitelisting is necessary for accurate evaluation. It's worth mentioning that sequence 6 has valid annotations for all objects in all frames.
  
    \subsection{Frame-drops of RGB-D camera}
    We encountered frame-drops of the Realsense camera. These frame-drops are reflected in the dataset by gaps in the RGB and depth image numbering and consequently the ground-truth poses.
    However, these frame drops accurately represent the camera's real-world behavior and do not pose a problem when using RGB(-D) data for pose estimation.

    When transferring the pose data to the event stream, special care must be taken to handle gaps between frames. Without explicit handling, adjacent frames become more than 33 ms apart, which is not addressed in the supplementary programs.

    \section{Dataset bias experiments}
    \label{sec:evaluation}
    
    In this section we evaluate pose-estimation algorithms on our novel dataset, focusing only on the RGB setting without further refinement.
    As the RCNN detector from \cite{labbe2020cosypose} failed on our dataset, we use the YOLOX detector as trained in \cite{liu2022gdrnpp_bop}.
    To ensure a fair comparison, we replicate the outcomes on the YCB-V dataset using YOLOX detections for all algorithms.

    It is worth highlighting that this initial evaluation does not capture all aspects of the dataset. Specifically, it does not include pose detection on the event data.

	\begin{table}
	\centering
	\begin{tabular}{|c|c|c|}
		\hline
		Algorithm & YCB-V & YCB-Ev (ours) \\
		\hline
   		\hline
		YOLOX \cite{ge2021yolox} & 89.8 & 50.1 \\
		\hline
		GDRNPP \cite{liu2022gdrnpp_bop} & 76.4 & 8.9 \\
		\hline
		Cosypose \cite{labbe2020cosypose} & 82.5 & 41.4 \\
		\hline
	\end{tabular}
	\caption{Average recall on RGB data. Computed as $AP_0$ for YOLOX detector and at a 2cm translation threshold for the pose estimation datasets. All models were trained on PBR+real.}
	\label{tbl:evaluation}
	\end{table}

	Table \ref{tbl:evaluation} shows the evaluation results for the 12 sequences that match between YCB-V and our dataset (refer to Table \ref{tbl:sequences}). The first row presents the average detection recall using YOLOX, representing the percentage of detected ground truth objects. A notable $\Delta$ 39.7 AR decrease in performance with YOLOX is observed, likely due to a dataset bias impacting the detector.

	Similar to BOP22, we calculate the average pose estimation recall for objects detected by YOLOX. For simplicity, we use a 2cm threshold on the translation vector, ignoring orientation and the need to account for object symmetry.
	
	The results in Table \ref{tbl:evaluation} demonstrate that we can replicate the average recall of "GDRNPP-PBRReal-RGB-SModel" and "CosyPose-ECCV20-SYNT+REAL-1VIEW," the top algorithms of 2022 and 2020, respectively. However, while the synthetic-to-real gap has been shown to reduce in the BOP22 \cite{sundermeyer2023bop} challenge, our results indicate that GDRNPP is actually more susceptible to dataset bias with a $\Delta$ 67.5 AR compared to CosyPose with a $\Delta$ 41.1 AR. This suggests that the reduction might be due to dataset-specific domain randomization on top of BlenderProc \cite{denninger2020blenderproc} rather than actual algorithmic improvements.

    \section{Conclusion and future work}
    \label{sec:conclusion}
    
    We have presented a benchmark dataset for 6DoF pose-estimation tasks on event vision data. Our capturing pipeline allows providing ground-truth poses even under fast camera motions.
    Our evaluation of dataset bias reveals that narrowing the domain gap is not enough to reduce the dataset bias to a satisfactory level.

    The main limitation of this work is the annotation accuracy. This is due to several sources of errors, including inaccuracies in the object models, inaccuracies in the synchronization between the event and color modalities, and uncertainties in the cameras intrinsic and extrinsic calibrations.
    
    In the future, problems with the RGB based annotations can be overcome by using pose estimation algorithms that operate directly on event data. This will allow for the annotation of challenging sequences using the event modality and the transfer of the estimated poses back to color images, enhancing the performance of RGB-only algorithms.
    
    Furthermore, we intend to add synthetic event data based on vid2e \cite{gehrig2020video} to overcome the annotation accuracy problems with simulation.

    \section*{Acknowledgements}
    This project was funded in part by the European Union’s Horizon Europe research and innovation programme under grant agreement No. 101120726.

    \bibliographystyle{splncs04}
    \bibliography{bibliography}
    \input{appendix}
\end{document}

%% file: appendix.tex
    \appendix
\section{Appendix}
\begin{table*}
    \centering
    \begin{tabular}{|l|c|c|c|c|c|c|c|}
        \hline
        Object & \multicolumn{2}{c|}{YOLOX} & \multicolumn{2}{c|}{GDRNPP}  & \multicolumn{2}{c|}{Cosypose} \\
        \hline
        \hline
        & YCB-V & YCB-Ev & YCB-V & YCB-Ev & YCB-V & YCB-Ev \\
\hline
002\_master\_chef\_can & 1.0 & 0.6 & 0.69 & 0.081 & 0.82 & 0.54 \\
\hline
003\_cracker\_box & 0.72 & 0.59 & 0.36 & 0.049 & 0.75 & 0.66 \\
\hline
004\_sugar\_box & 0.98 & 0.44 & 0.94 & 0.035 & 0.95 & 0.55 \\
\hline
005\_tomato\_soup\_can & 0.98 & 0.68 & 0.93 & 0.051 & 0.92 & 0.41 \\
\hline
006\_mustard\_bottle & 1.0 & 0.85 & 0.99 & 0.13 & 1.0 & 0.45 \\
\hline
007\_tuna\_fish\_can & 1.0 & 0.63 & 0.98 & 0.069 & 0.99 & 0.34 \\
\hline
008\_pudding\_box & 1.0 & 0.47 & 0.55 & 0.091 & 0.62 & 0.2 \\
\hline
009\_gelatin\_box & 0.73 & 0.42 & 0.9 & 0.02 & 0.9 & 0.29 \\
\hline
010\_potted\_meat\_can & 0.88 & 0.61 & 0.86 & 0.05 & 0.77 & 0.56 \\
\hline
011\_banana & 1.0 & 0.43 & 0.67 & 0.26 & 0.84 & 0.085 \\
\hline
019\_pitcher\_base & 1.0 & 0.92 & 0.97 & 0.089 & 0.97 & 0.41 \\
\hline
021\_bleach\_cleanser & 0.81 & 0.87 & 0.83 & 0.18 & 0.87 & 0.34 \\
\hline
024\_bowl & 0.61 & 0.4 & 0.29 & 0.1 & 0.44 & 0.79 \\
\hline
025\_mug & 1.0 & 0.41 & 0.58 & 0.005 & 0.9 & 0.76 \\
\hline
035\_power\_drill & 1.0 & 0.65 & 0.88 & 0.063 & 0.9 & 0.45 \\
\hline
036\_wood\_block & 0.99 & 0.53 & 0.32 & 0.097 & 0.25 & 0.56 \\
\hline
037\_scissors & 0.81 & 0.8 & 0.64 & 0.16 & 0.14 & 0.57 \\
\hline
040\_large\_marker & 1.0 & 0.74 & 0.64 & 0.14 & 0.8 & 0.43 \\
\hline
051\_large\_clamp & 0.26 & 0.53 & 0.37 & 0.04 & 0.72 & 0.43 \\
\hline
052\_extra\_large\_clamp & 0.82 & 0.7 & 0.25 & 0.15 & 0.36 & 0.57 \\
\hline
061\_foam\_brick & 1.0 & 0.61 & 0.69 & 0.051 & 0.78 & 0.49 \\
        \hline
    \end{tabular}
    \caption{Per object results for \autoref{tbl:evaluation}}
\end{table*}

\begin{figure*}
    \includegraphics[width=0.33\textwidth]{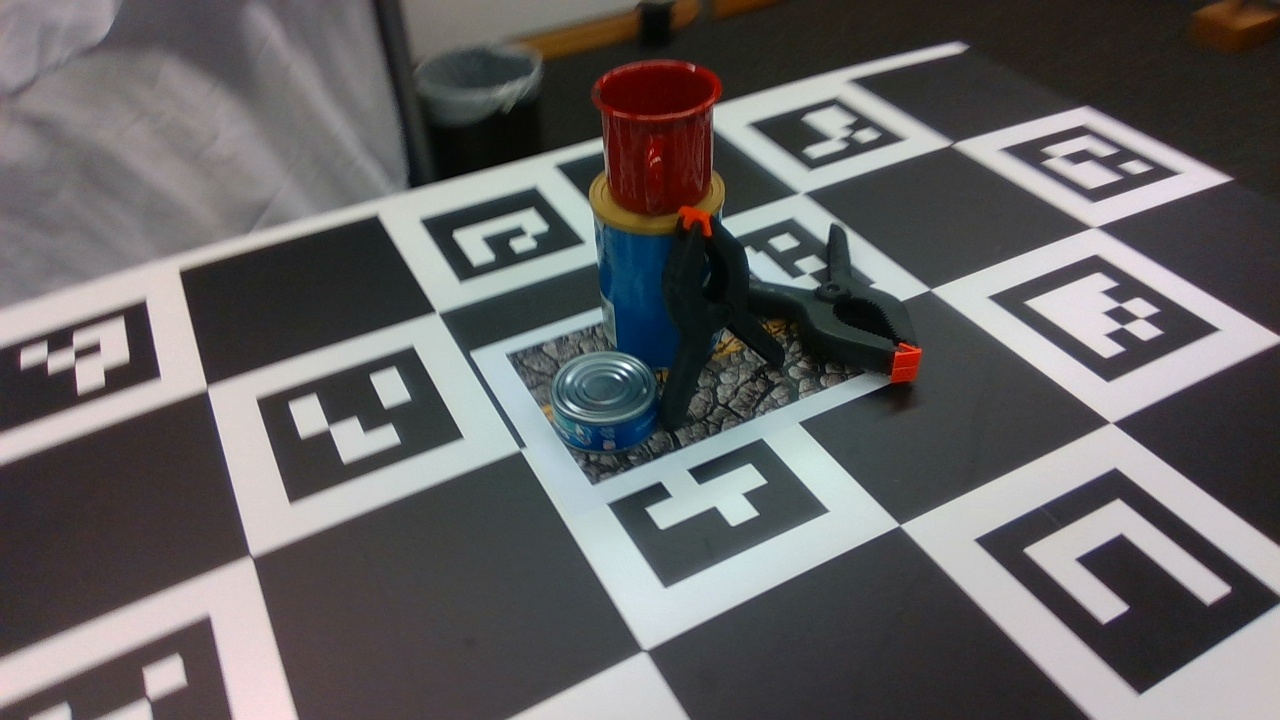}
    \includegraphics[width=0.33\textwidth]{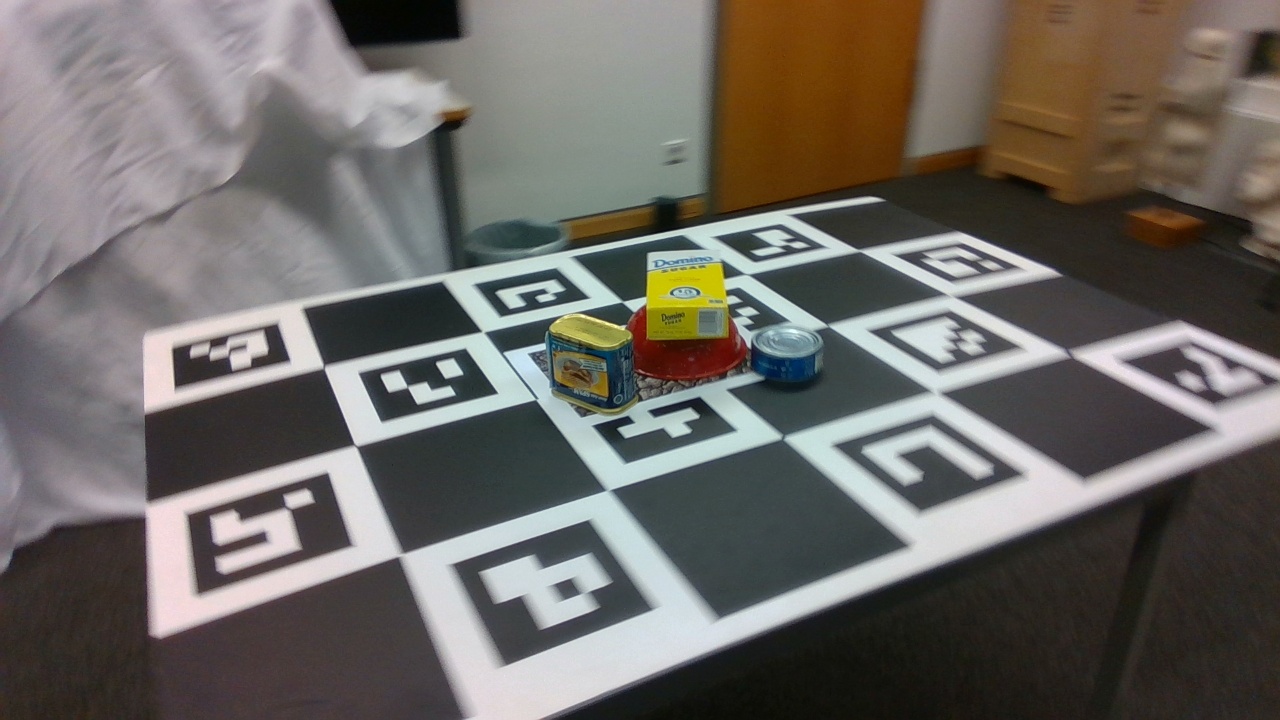}
    \includegraphics[width=0.33\textwidth]{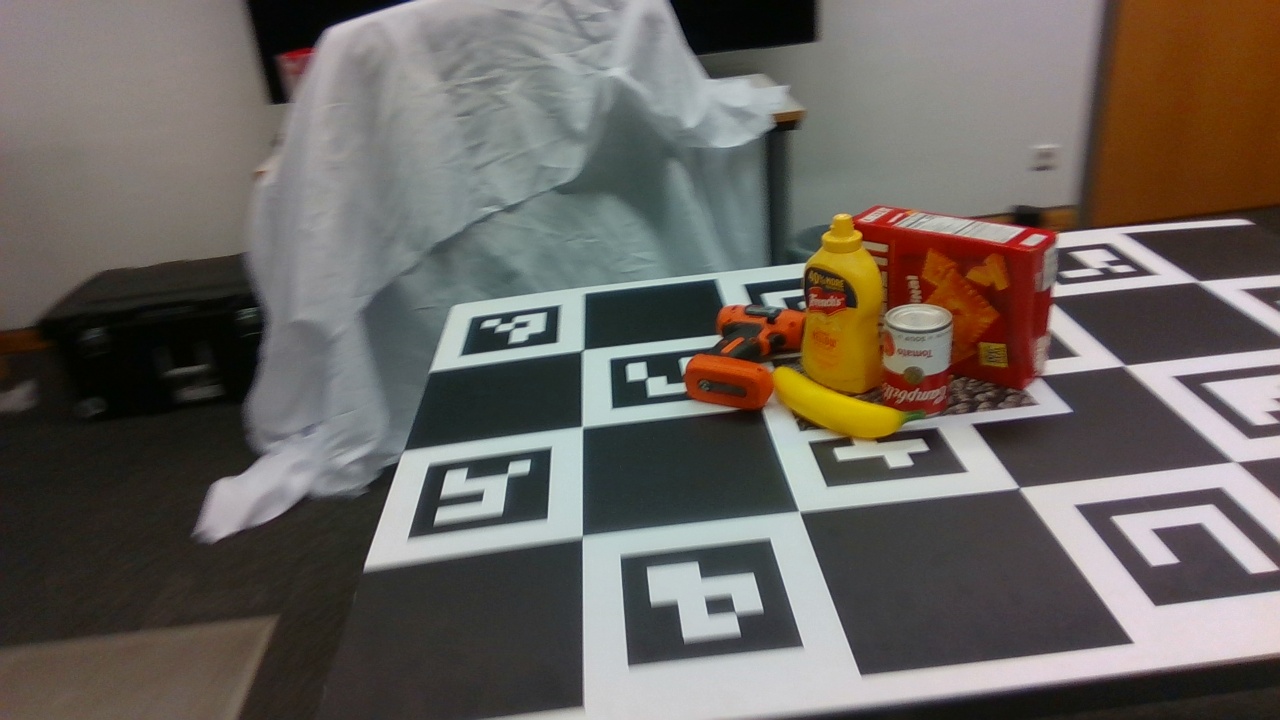}
    \includegraphics[width=0.33\textwidth]{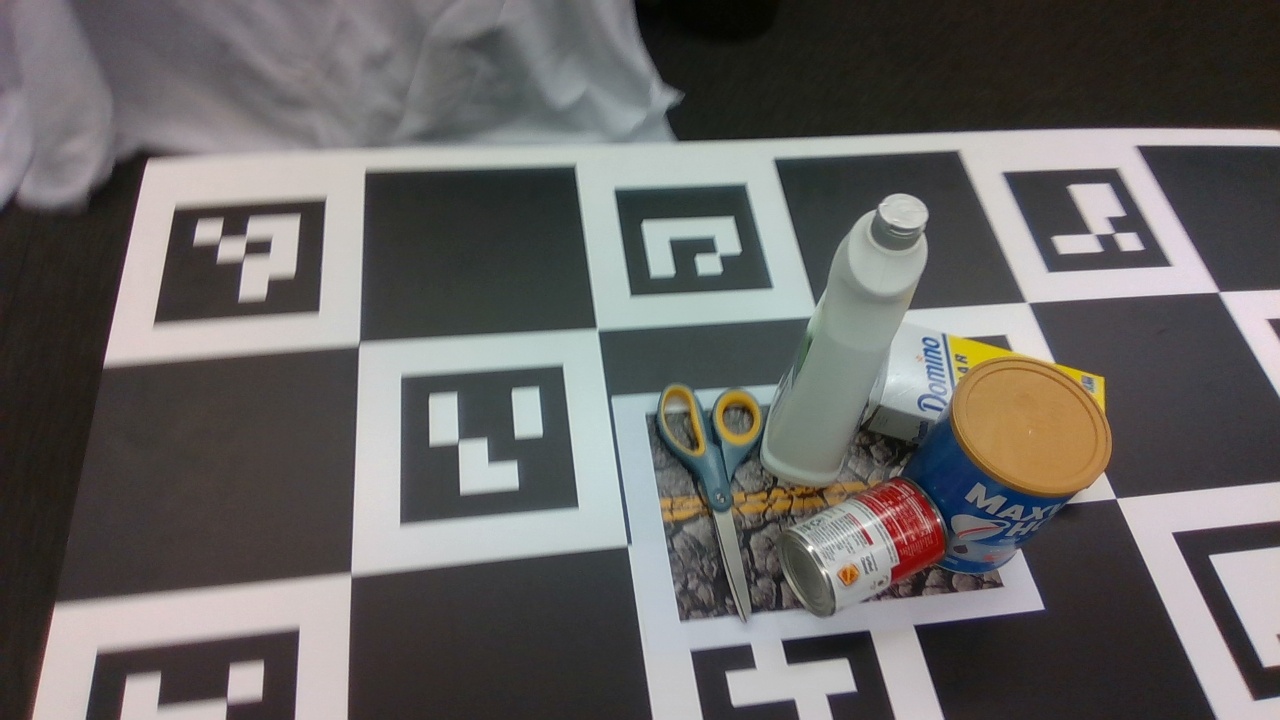}
    \includegraphics[width=0.33\textwidth]{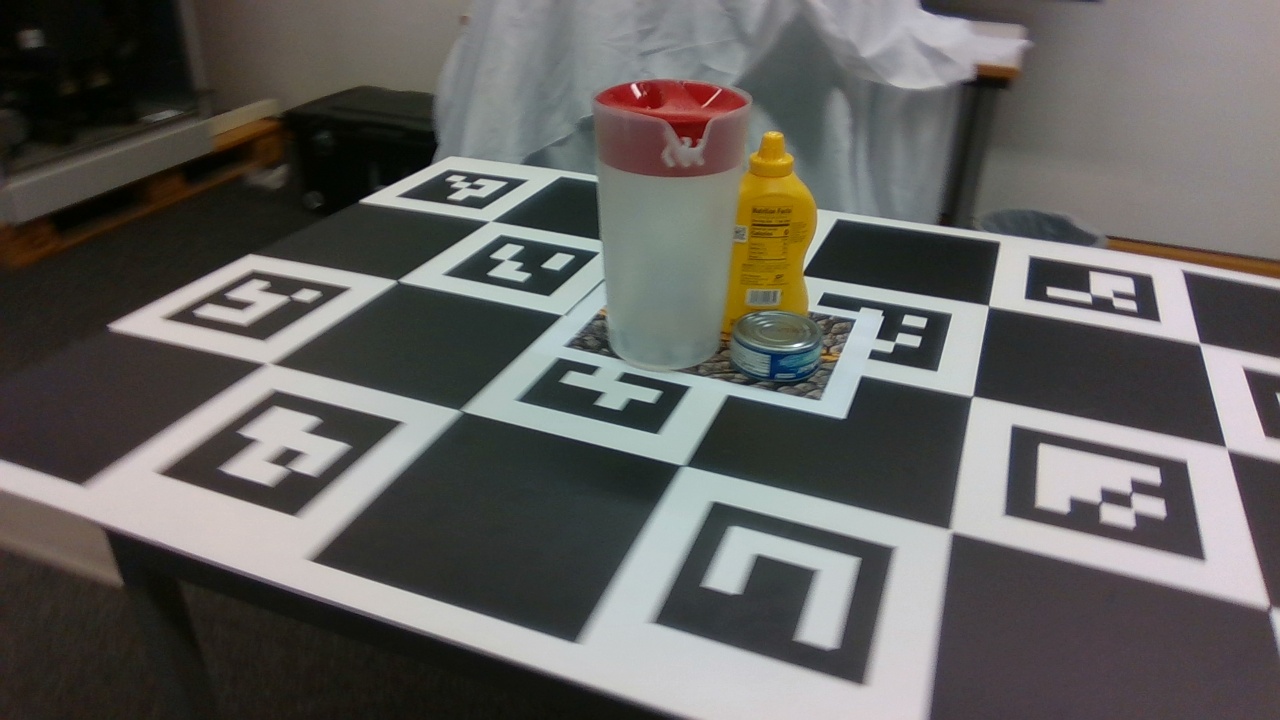}
    \includegraphics[width=0.33\textwidth]{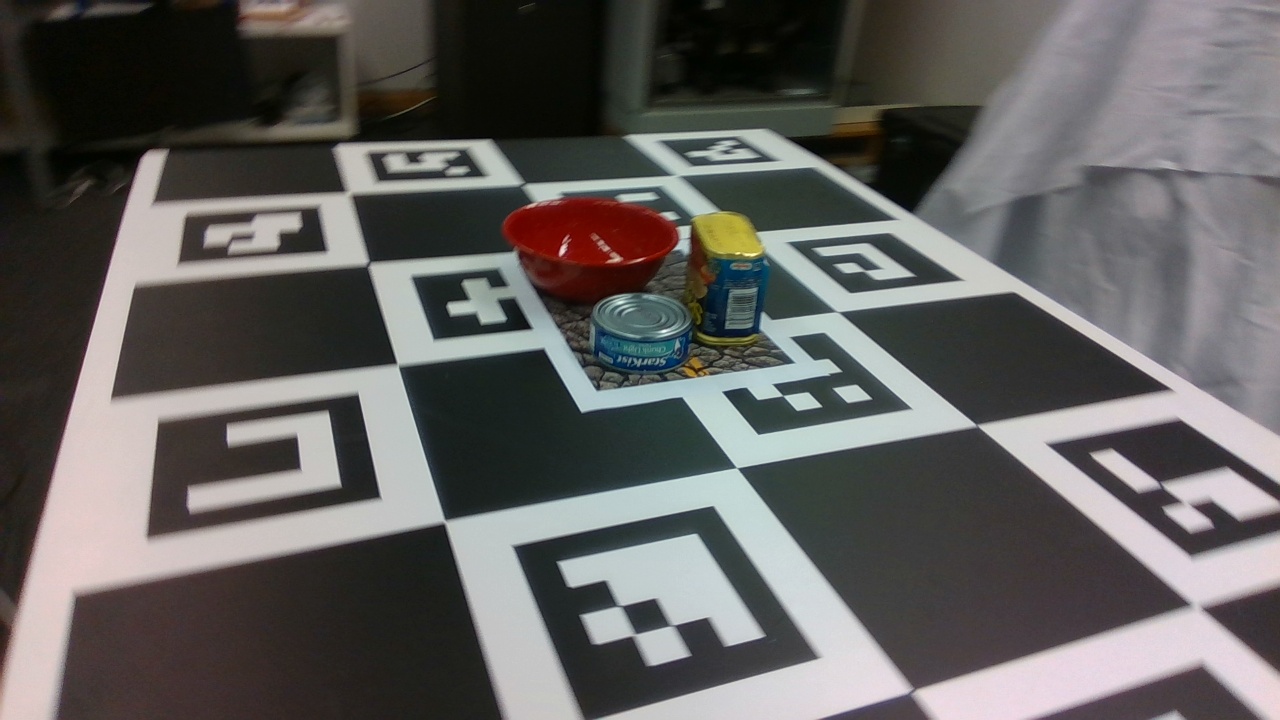}
    \includegraphics[width=0.33\textwidth]{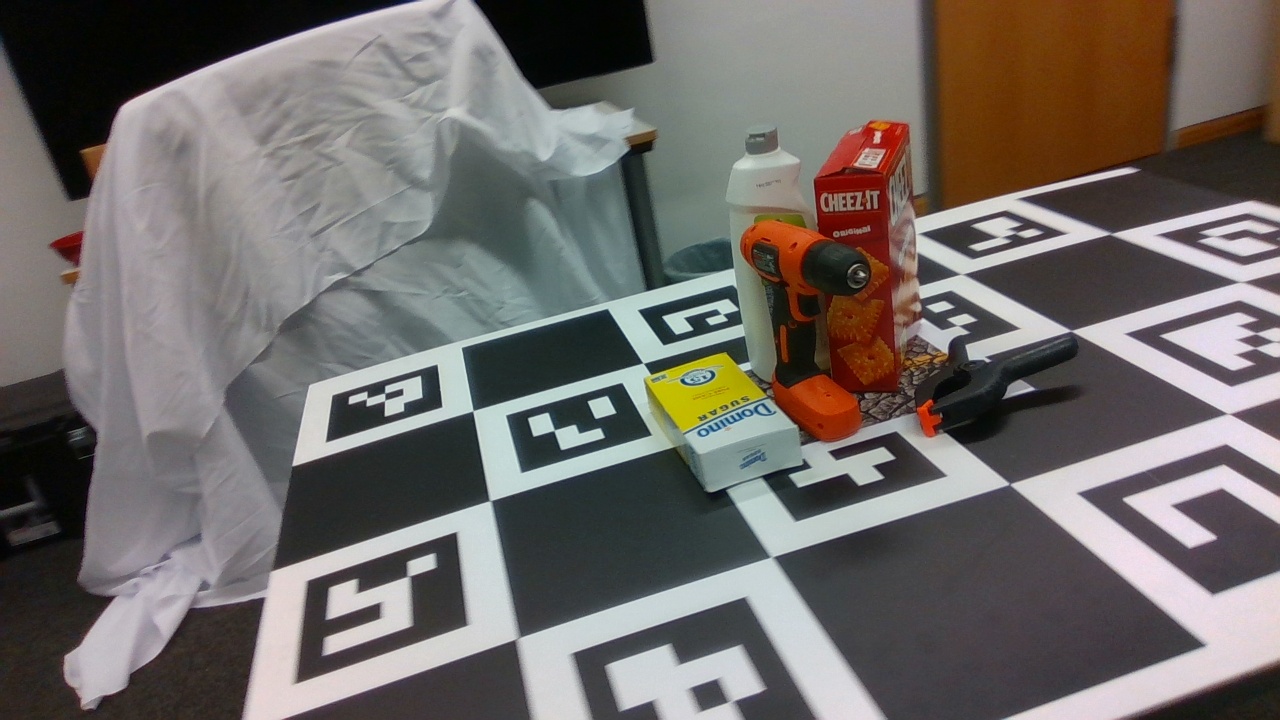}
    \includegraphics[width=0.33\textwidth]{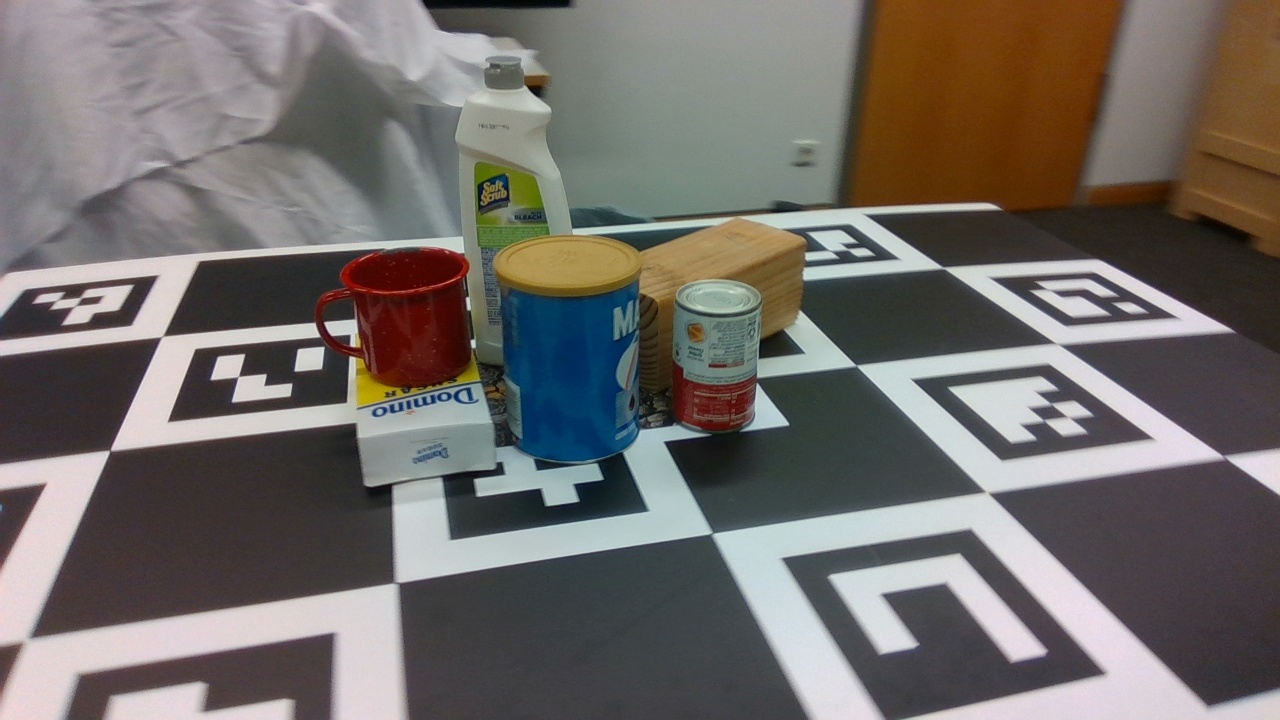}
    \includegraphics[width=0.33\textwidth]{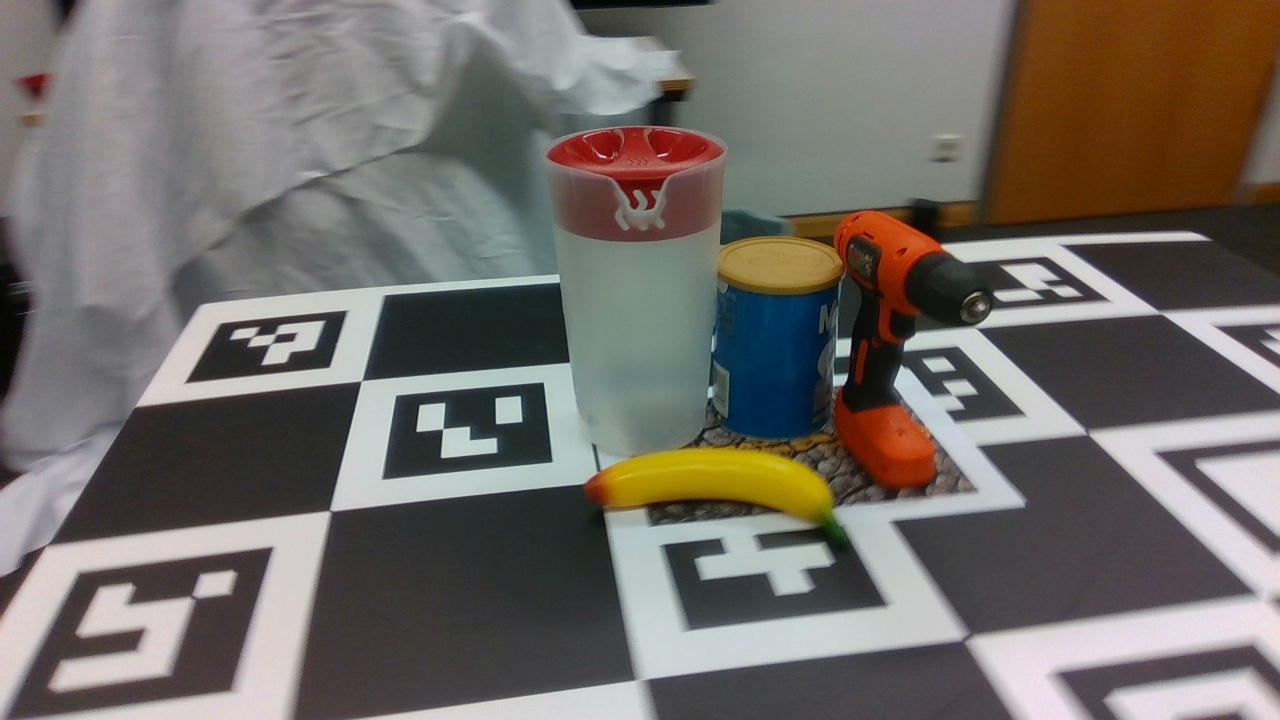}
    \includegraphics[width=0.33\textwidth]{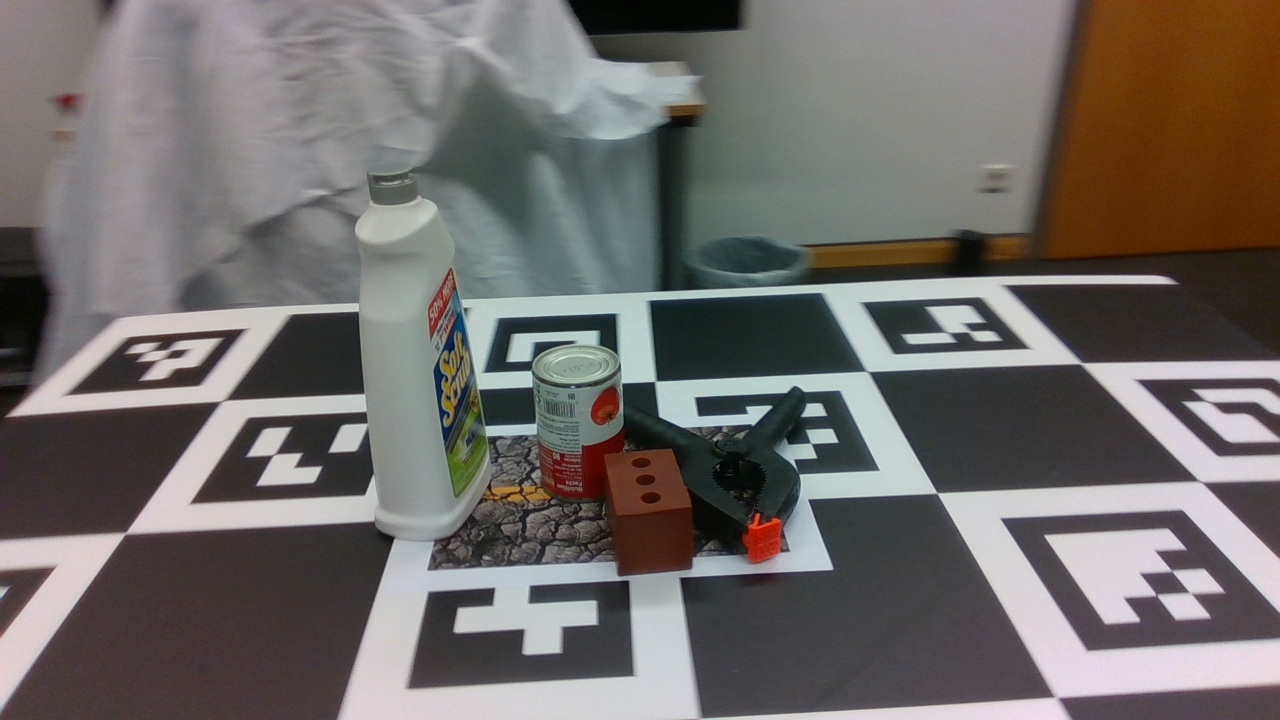}
    \includegraphics[width=0.33\textwidth]{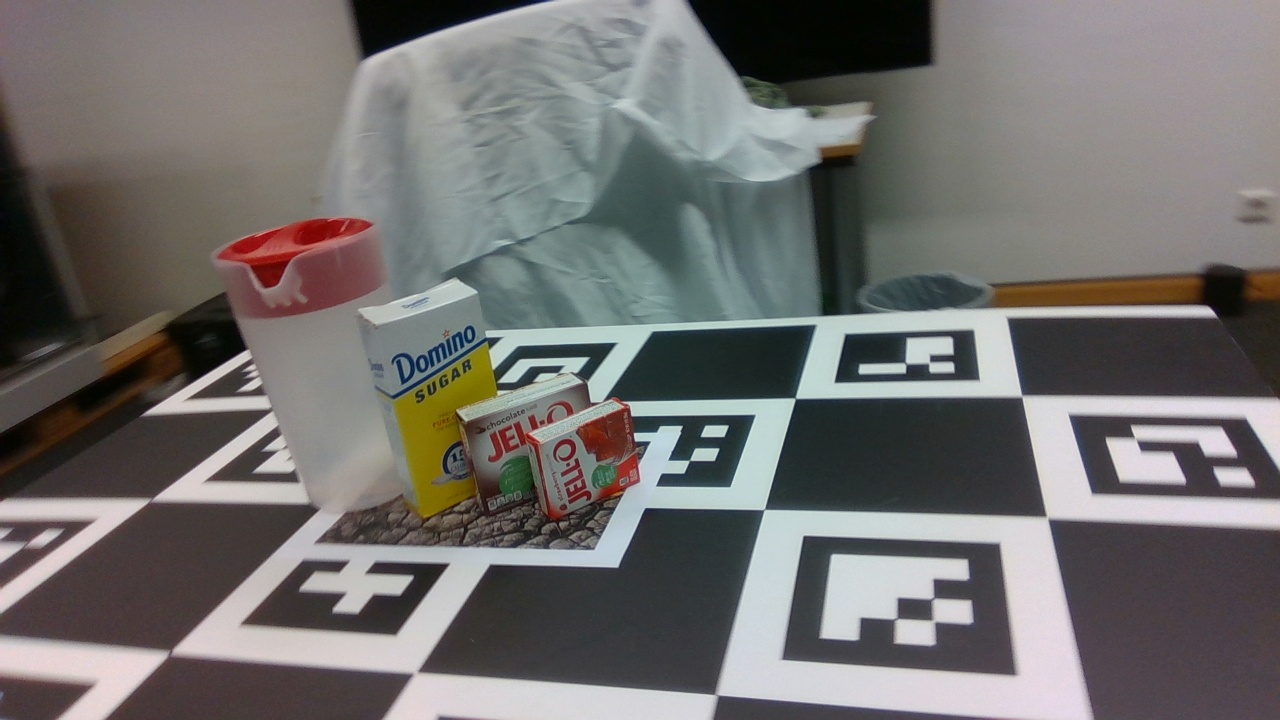}
    \includegraphics[width=0.33\textwidth]{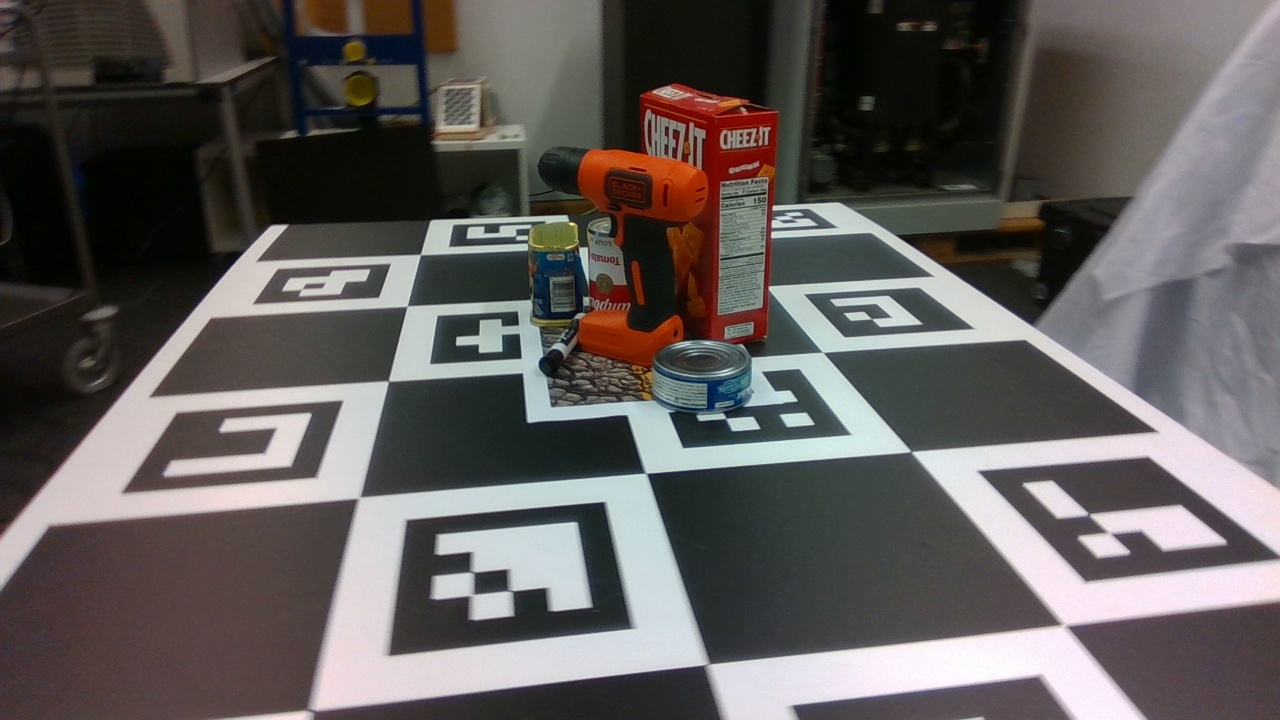}
    \includegraphics[width=0.33\textwidth]{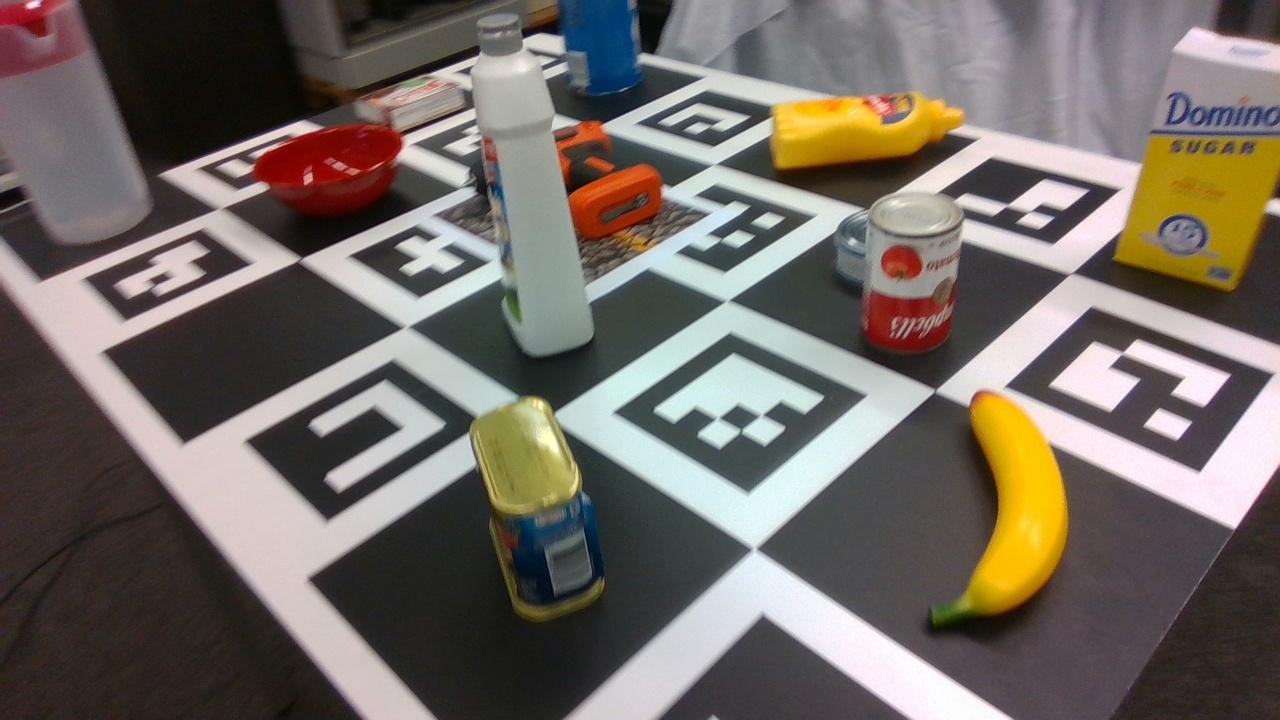}
    \includegraphics[width=0.33\textwidth]{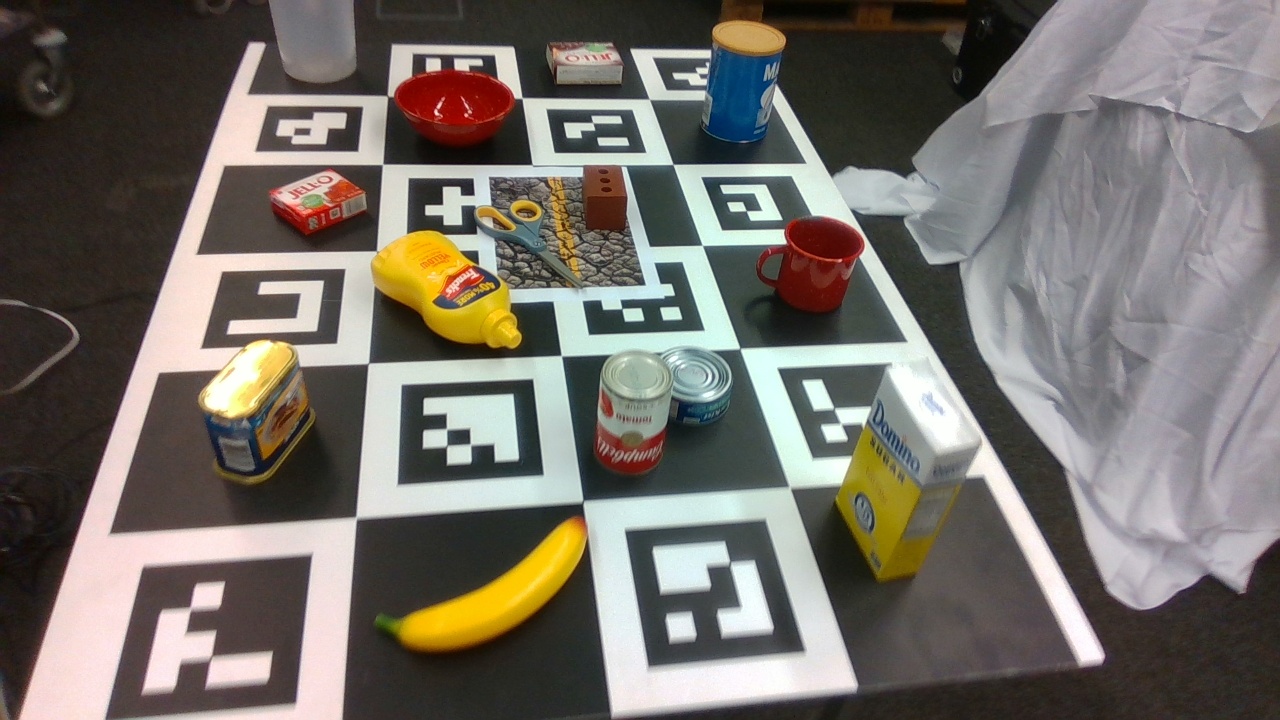}
    \includegraphics[width=0.33\textwidth]{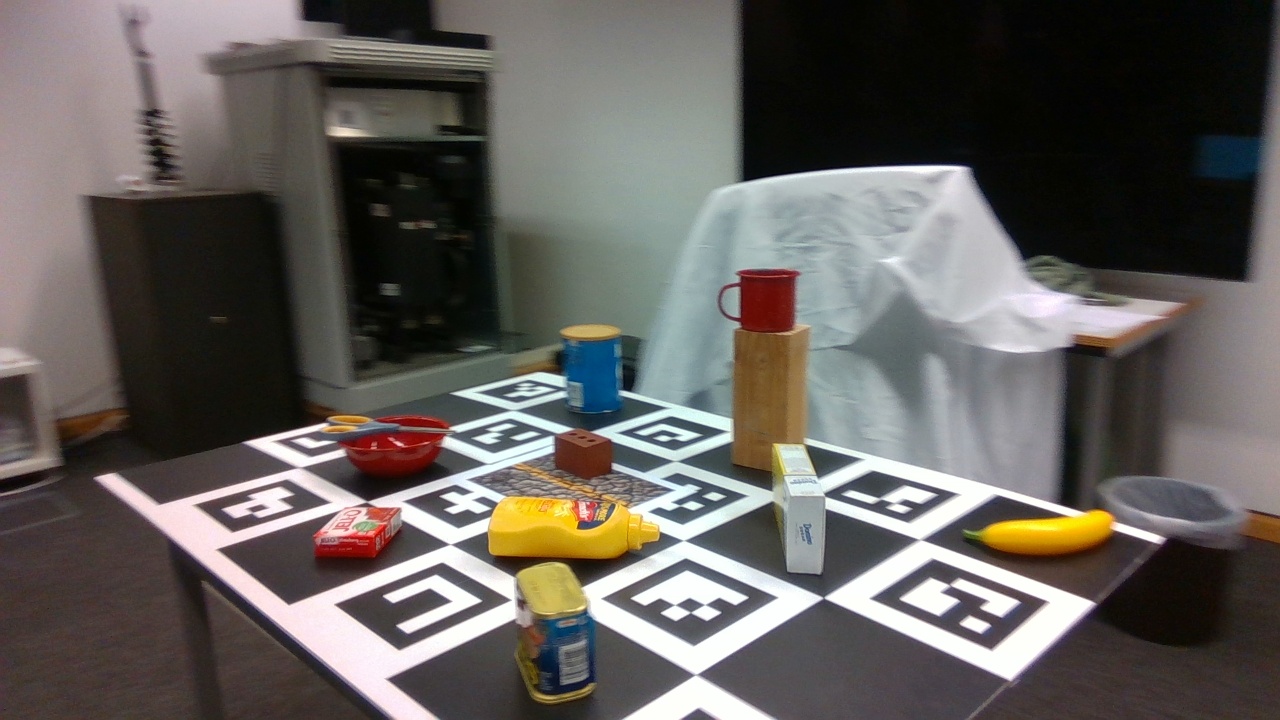}
    \includegraphics[width=0.33\textwidth]{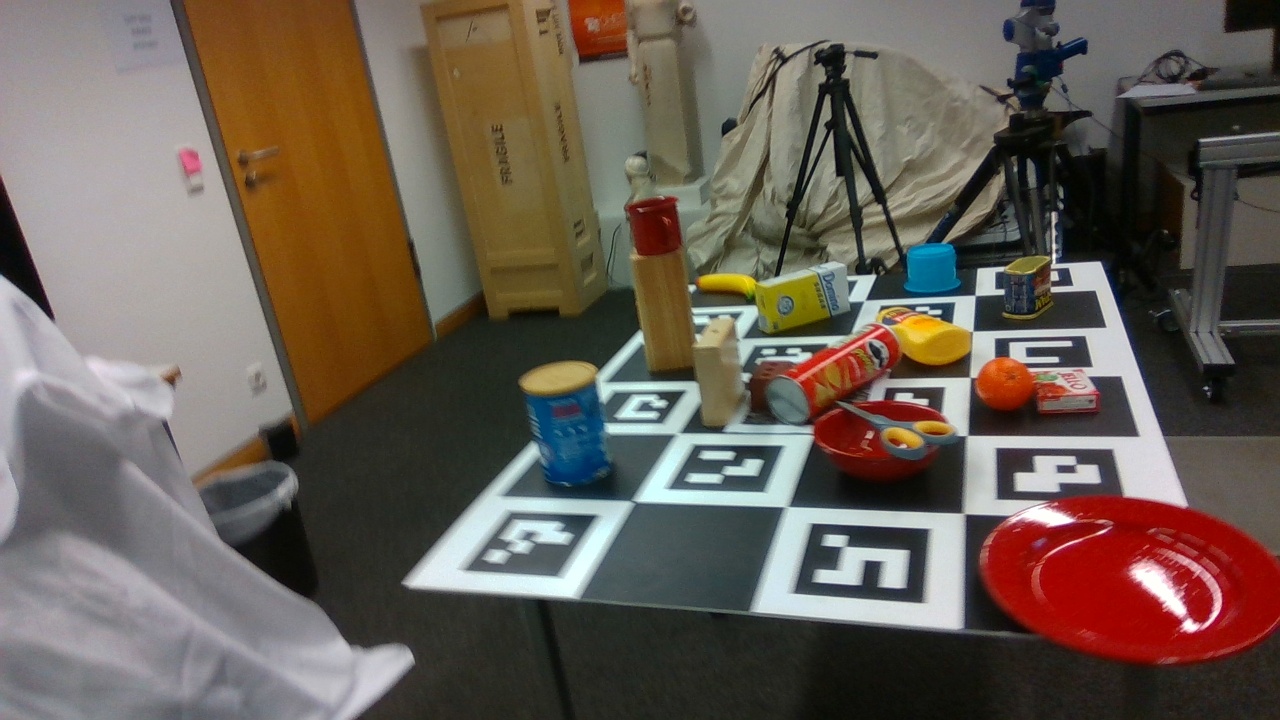}
    \includegraphics[width=0.33\textwidth]{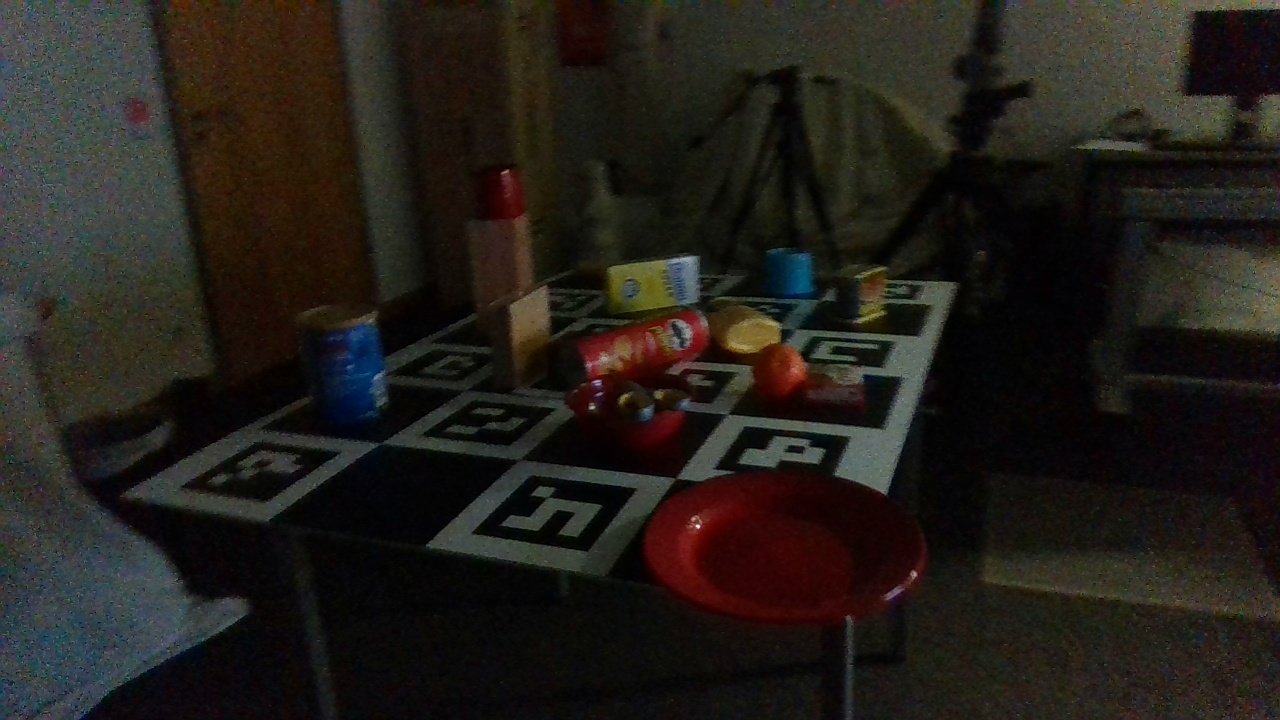}
    \includegraphics[width=0.33\textwidth]{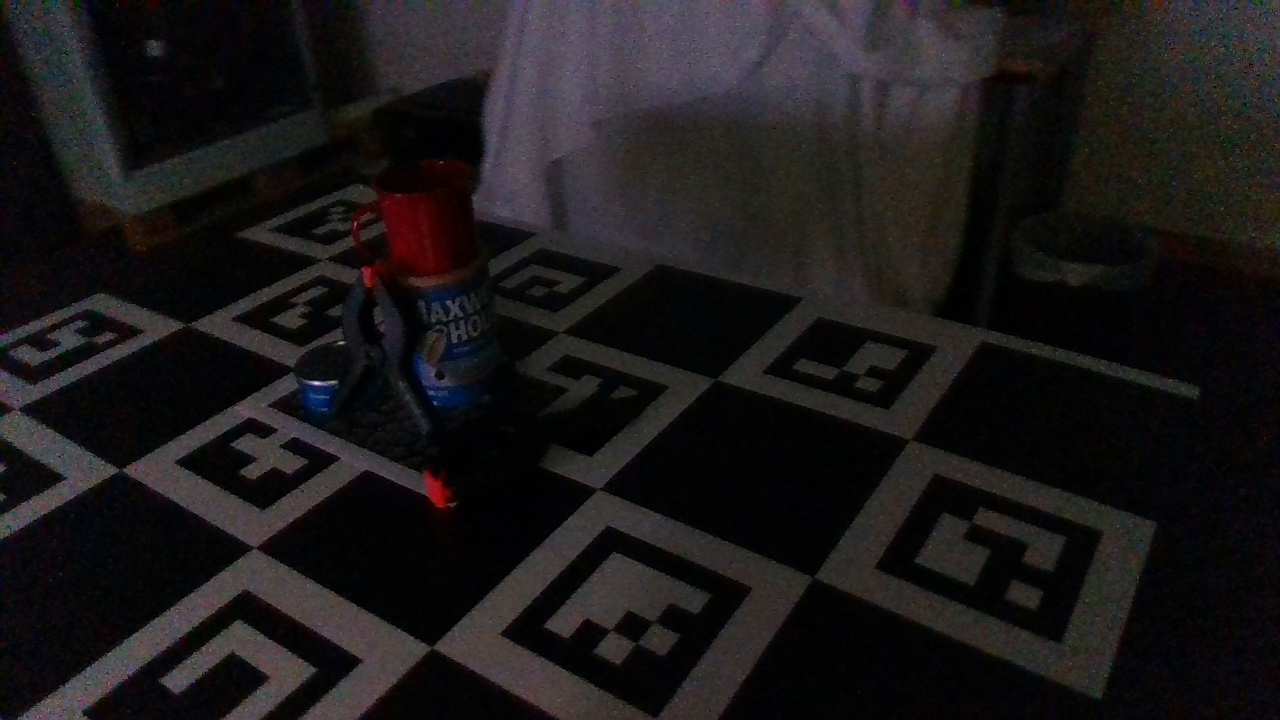}
    \includegraphics[width=0.33\textwidth]{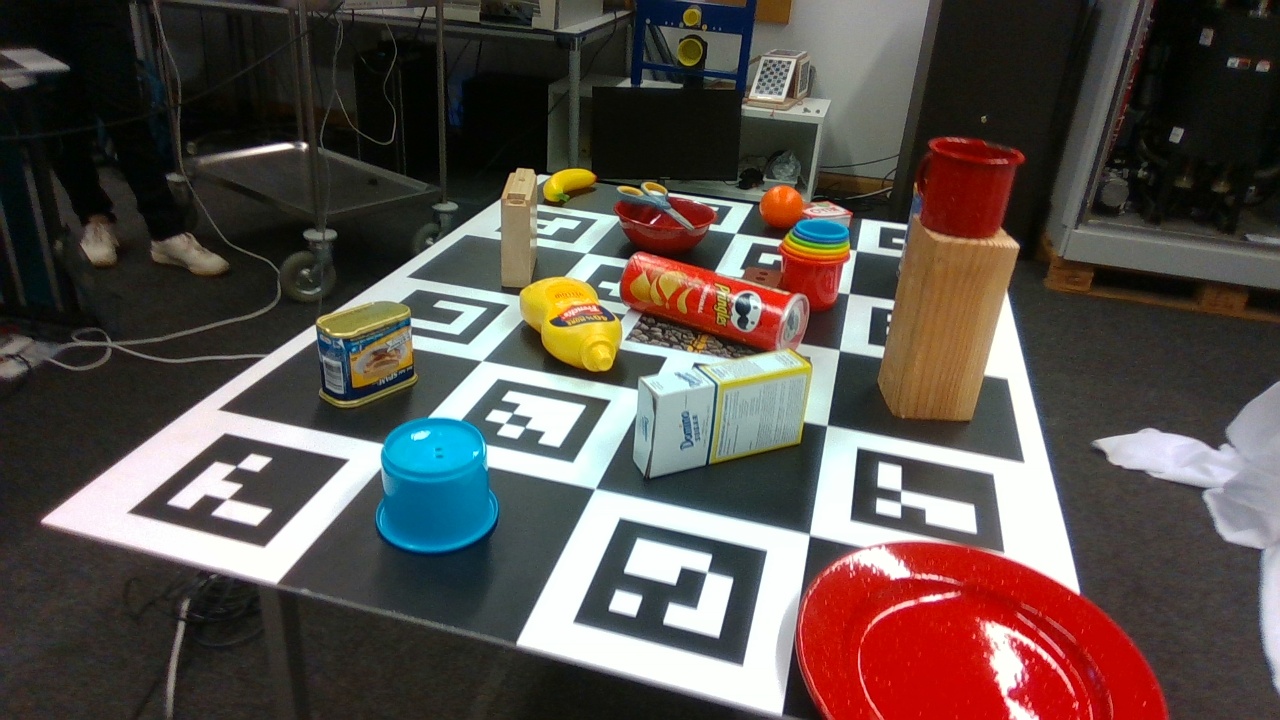}
    \includegraphics[width=0.33\textwidth]{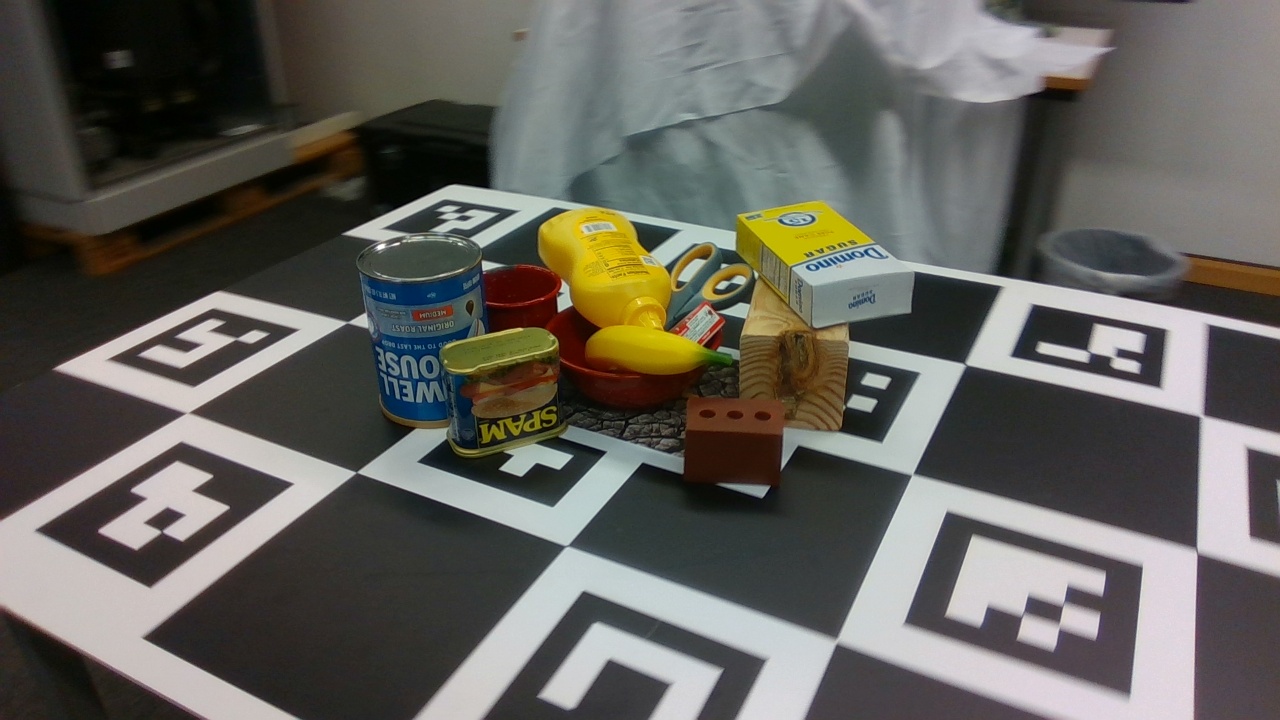}
    \includegraphics[width=0.33\textwidth]{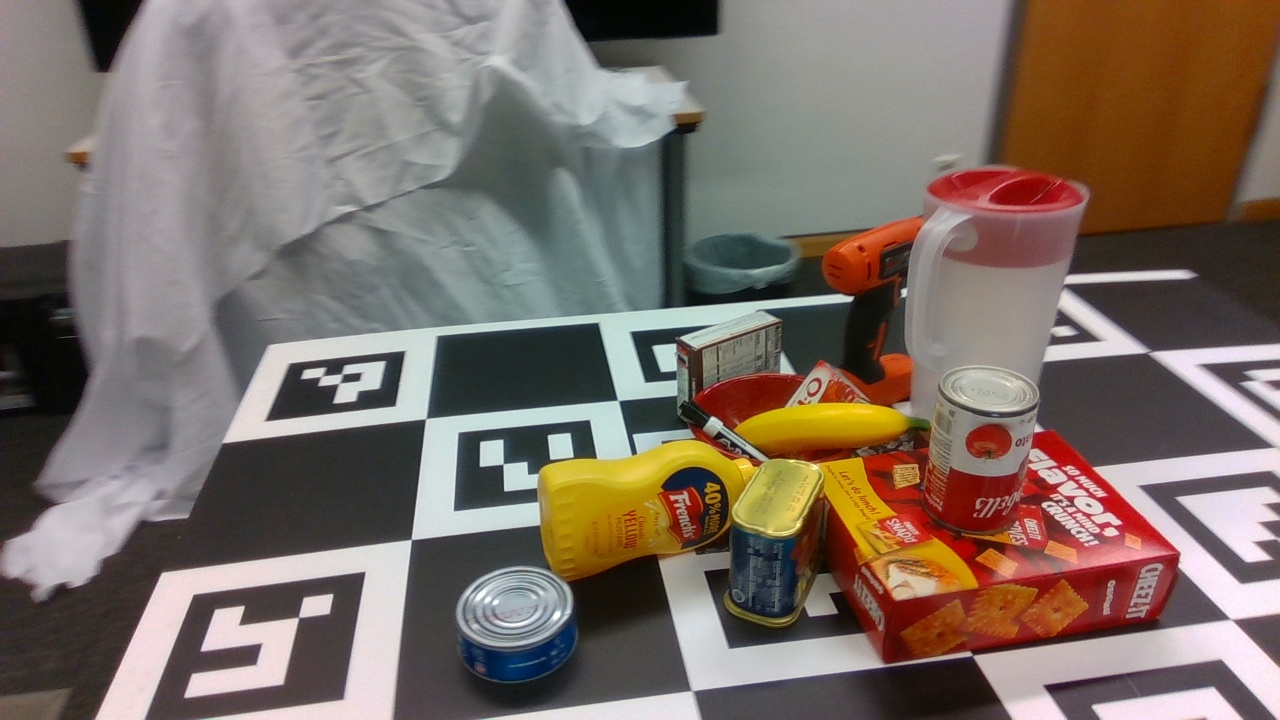}
    \caption{The 21 sequences in our dataset.}
    \label{fig:seq_overview}
\end{figure*}